\documentclass[conference]{IEEEtran}
\IEEEoverridecommandlockouts
\usepackage{cite}
\usepackage[a4paper, total={6in, 8in}]{geometry}
\usepackage{amsmath,amssymb,amsfonts}
\usepackage{algorithmic}
\usepackage{graphicx}
\usepackage{textcomp}
\usepackage{xcolor}
\usepackage{subcaption}
\def\BibTeX{{\rm B\kern-.05em{\sc i\kern-.04em b}\kern-.20em
    T\kern-.1667em\lower.7ex\hbox{E}\kern-.125emX}}
\begin{document}
\title{Multimodal Speech Emotion Recognition and Ambiguity Resolution\\
{\footnotesize}
\thanks{\textsuperscript{1}Code for all the experiments is available at \texttt{http://tinyurl.com/y55dlc3m}}
}

\author{\IEEEauthorblockN{Gaurav Sahu}
\IEEEauthorblockA{\textit{David R. Cheriton School of Computer Science} \\
\textit{University of Waterloo}\\
Ontario, Canada \\
\texttt{gaurav.sahu@uwaterloo.ca}}
}

\maketitle

\begin{abstract}
Identifying emotion from speech is a non-trivial task pertaining to the ambiguous definition of emotion itself. In this work, we adopt a feature-engineering based approach to tackle the task of speech emotion recognition. Formalizing our problem as a multi-class classification problem, we compare the performance of two categories of models. For both, we extract eight hand-crafted features from the audio signal. In the first approach, the extracted features are used to train six traditional machine learning classifiers, whereas the second approach is based on deep learning wherein a baseline feed-forward neural network and an LSTM-based classifier are trained over the same features. In order to resolve ambiguity in communication, we also include features from the text domain. We report accuracy, f-score, precision and recall for the different experiment settings we evaluated our models in. Overall, we show that lighter machine learning based models trained over a few hand-crafted features are able to achieve performance comparable to the current deep learning based state-of-the-art method for emotion recognition.
\end{abstract}


\section{Introduction}
Communication is the key to human existence and more often than not, we have to deal with ambiguous situations. For instance, the phrase ``This is awesome" could be said under either happy or sad settings. Humans are able to resolve ambiguity in most cases because we can efficiently comprehend information from multiple domains (henceforth, referred to as modalities), namely, speech, text and visual. With the rise of deep learning algorithms, there have been multiple attempts to tackle the task of Speech Emotion Recognition (SER) as in \cite{huang2014speech} \cite{yoon2018multimodal} and \cite{han2014speech}. However, this rise has made practitioners rely more on the power of the deep learning models as opposed to using domain knowledge to construct meaningful features and building models that perform well as well as are interpretable. In this work, we explore the implication of hand-crafted features for SER and compare the performance of lighter machine learning models with the heavily data-reliant deep learning models. Furthermore, we also combine features from the textual modality to understand the correlation between different modalities and aid ambiguity resolution. More formally, we pose our task as a multi-class classification problem and employ the two classes of models to solve that. For both the approaches, we first extract hand-crafted features from the time domain of the audio signal and train the respective models.

In the first approach, we train traditional machine learning classifiers, namely, Random Forests, Gradient Boosting, Support Vector Machines, Naive-Bayes and Logistic Regression. In the second approach, we build a Multi-Layer Perceptron and an LSTM \cite{hochreiter1997long} classifier to recognize emotion given a speech signal. The models are evaluated on the \textit{IEMOCAP} \cite{busso2008iemocap} dataset under different settings, namely, \textit{Audio-only}, \textit{Text-only} and \textit{Audio + Text} \textsuperscript{1}.

The rest of the paper is organized as follows: Section II describes existing methods in the literature for the task of speech emotion recognition; Section III gives an overview of the dataset used in this work and the pre-processing steps applied before feature extraction; Section IV describes the proposed models and implementation details; Results are reported in Section V, followed by the conclusion and future scope of this work in Section VI.

\section{Literature Review}
In this section, we review some of the work that has been done in the field of speech emotion recognition (SER). The task of SER is not new and has been studied for quite some time in literature. A majority of the early approaches (\cite{nogueiras2001speech} \cite{schuller2003hidden}) used Hidden Markov Models (HMMs) \cite{rabiner1986introduction} for identifying emotion from speech. Recent introduction of deep neural networks to the domain has also significantly improved the state-of-the-art performance. For instance, \cite{han2014speech} and \cite{kim2013deep} use recurrent autoencoders to solve the task. Recently, methods have also been proposed to efficiently combine features from multiple domains, such as, Tensor Fusion Networks \cite{zadeh2017tensor} and Low-Rank Matrix Multiplication \cite{liu2018efficient}, instead of trivial concatenation.

This work aims to provide a comparative study between 1) deep learning based models that are trained end-to-end, and 2) lighter machine learning and deep learning based models trained over hand-crafted features. We also investigate the information residing in multiple modalities and how their combination affects the performance.

\section{Dataset}
In this work, we use the \textit{IEMOCAP} \cite{busso2008iemocap} released in 2008 by researchers at the University of Southern California (USC). It contains five recorded sessions of conversations from ten speakers and amounts to nearly 12 hours of audio-visual information along with transcriptions. It is annotated with eight categorical emotion labels, namely, anger, happiness, sadness, neutral, surprise, fear, frustration and excited. It also contains dimensional labels such as values of the activation and valence from 1 to 5; however, they are not used in this work.

The dataset is already split into multiple utterances for each session and we further split each utterance file to obtain \texttt{wav} files for each sentence. This was done using the start timestamp and end timestamp provided for the transcribed sentences. This results in a total of $\sim$10K audio files which are then used to extract features.

\section{Methodology}
This section describes the data pre-processing steps followed by a detailed description of the features extracted and the two models applied to the classification problem.
\subsection{Data Pre-processing}
\paragraph{Audio} A preliminary frequency analysis revealed that the dataset is not balanced. The emotions ``fear" and ``surprise" were under-represented and use upsampling techniques to alleviate the issue. We then merged examples from ``happy" and ``excited" classes as ``happy"  was under-represented and the two emotions closely resemble each other. In addition to that, we discard examples classified as ``others"; they corresponded to examples that were labeled ambiguous even for a human. Applying the aforementioned operations resulted in 7837 examples in total. Final sample distribution for each of the emotions is shown in Table \ref{tab:emo_freq}.

\begin{table}[]
    \centering
    \caption{Number of examples for each emotion}
    \begin{tabular}{|c|c|}
        \hline
        \textbf{Class} & \textbf{Count} \\
        \hline
        Angry & 860 \\
        \hline
        Happy & 1309 \\
        \hline
        Sad & 2327 \\
        \hline
        Fear & 1007 \\
        \hline
        Surprise & 949 \\
        \hline
        Neutral & 1385 \\
        \hline
        \textbf{Total} & 7837 \\
        \hline
    \end{tabular}
    \label{tab:emo_freq}
\end{table}

\paragraph{Text} The available transcriptions were first normalized to lowercase and any special symbols were removed.

\subsection{Feature Extraction} We now describe the handcrafted features used to train both, the ML- and the DL-based models.
\subsubsection{Audio Features}
\paragraph{Pitch} Pitch is important because waveforms produced by our vocal cords change depending on our emotion. Many algorithms for estimating the pitch signal exist. We use the most common method based on \textit{autocorrelation of center-clipped} frames \cite{sondhi1968new}. Formally, the input signal $y[n]$ is center-clipped to give a resultant signal, $y_{clipped}[n]$:

\begin{equation}
    y_{clipped}[n]= 
    \begin{cases}
    y[n] - C_l,& \text{if } y[n]\geq C_l\\
    0,              & \text{if } |y[n]| < C_l\\
    y[n] + C_l, & \text{if } y[n] \leq C_l
    \end{cases}
\end{equation}

Typically, $C_l$ is nearly half the mean of the input signal and $[\cdot]$ denotes the discrete nature of the input signal. Now, autocorrelation is calculated for the obtained signal $y_{clipped}$, which is further normalized and the peak values associated with the pitch of the given input $y[n]$. It was found that center-clipping the input signal resulted in more distinct autocorrelation peaks.

\begin{figure}[h!]
	\centering
	\includegraphics[width=\linewidth]{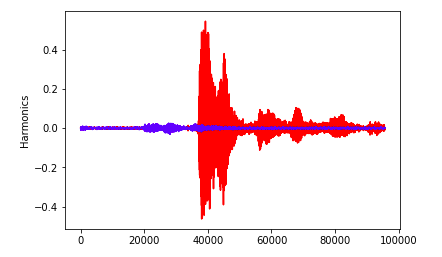}
	\caption{Harmonics of angry (red) and sad (blue) audio signals}
	\label{fig:harmonic}
\end{figure}

\paragraph{Harmonics} In the emotional state of anger or for stressed speech, there are additional excitation signals other than pitch (\cite{teager1990evidence}, \cite{zhou2001nonlinear}). This additional excitation is apparent in the spectrum as harmonics (see Figure \ref{fig:harmonic}) and cross-harmonics. We calculate harmonics using a median-based filter as described in \cite{fitzgerald2010harmonic}. First, the median filter is created for a given window size $l$, given by:

\begin{equation}
y[n] = median(x[n - k:n + k] | k = (l - 1)/2)
\label{eq:med_filter_l_odd}
\end{equation}

where $l$ is odd. For cases when $l$ is even, the median is obtained as the mean of two values in the middle of the sorted list. This filter is then applied to $S_h$, the $h-$th frequency slice of a given spectrogram $\boldsymbol{S}$, to get harmonic-enhanced spectrogram frequency slice $H_h$ as:

\begin{equation}
    H_i = M(S_h, l_{harm})
\end{equation}

Here $M$ is the median filter, $i$ is the $i-$th time step and $l_{harm}$ is the length of the harmonic filter.

\begin{figure}[h!]
    \centering
	\includegraphics[width=\linewidth]{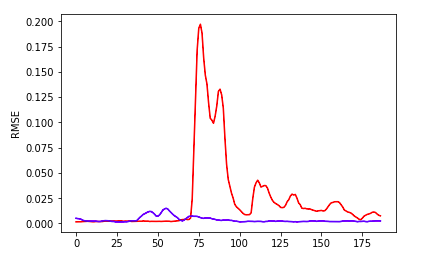}
	\caption{RMSE plots of angry (red) and sad (blue) audio signals}
	\label{fig:rmse}
\end{figure}

\paragraph{Speech Energy} Since the energy of a speech signal can be related to its loudness, we can use it to detect certain emotions. Figure \ref{fig:rmse} shows the difference in energy levels of an ``angry" signal v/s that of a ``sad" signal. We use standard Root Mean Square Energy (RMSE) to represent speech energy using the equation:

\begin{equation}
    E = \sqrt{\frac{1}{n}\sum_{i=1}^{n} y[i]^{2}}
    \label{eq:rmse}
\end{equation}

RMSE is calculated frame by frame and we take both, the average and standard deviation as features.

\paragraph{Pause} We use this feature to represent the ``silent" portion in the audio signal. This quantity is directly related to our emotions; for instance, we tend to speak very fast when excited (say, angry or happy, resulting in a low \textit{Pause} value). The feature value is given by:

\begin{equation}
    Pause = Pr(y[n] < t)
    \label{eq:pause}
\end{equation}

where $t$ represents a carefully-chosen threshold of $\approx 0.4*E$, $E$ being the RMSE.

\paragraph{Central moments} Finally, we use the mean and standard deviation of the amplitude of the signal to incorporate a ``summarized" information of the input.

\subsubsection{Text Features:}
\paragraph{Term Frequency-Inverse Document Frequency (TFIDF)} TFIDF is a numerical statistic that shows the correlation between a word and a document in a collection or corpus. It consists of two parts:

\begin{itemize}
    \item \textit{Term Frequency:} It denotes how many times a word/token occurs in a document. The simplest choice is to use \textit{raw count} of a token in a document (sentences, in our case).
    \item \textit{Inverse Document Frequency:} This term is introduced to lessen the bias due to frequently occurring words in language such ``the", ``a" and ``an". Usually, $idf$ for a term $t$ and a document $D$ is defined as:
    \begin{equation}
        idf(t, D) = \log \frac{N}{|d \in D: t \in d|}
    \end{equation}
    The denominator shows the frequency of documents containing the term $t$ and $N$ is the total number of documents.
\end{itemize}

Finally, TFIDF value for a term is calculated by taking the product of TF and IDF values.

\subsection{Machine Learning Models:} This section describes the various ML-based classifiers considered in this work, namely, Random Forests, Gradient Boosting, Support Vector Machines, Naive-Bayes, and Logistic Regression
\paragraph{Random Forest (RF)} Random forests are ensemble learners that operate by constructing multiple decision trees at training time and outputting the class that is the mode of the classes (classification) of the individual trees. It has two base working principles:

\begin{itemize}
    \item Each decision tree predicts using a random subset of features \cite{amit1997joint}
    \item Each decision tree is trained with only a subset of training samples. This is known as bootstrap aggregating \cite{breiman1996bagging}
\end{itemize}

Finally, a majority vote of all the decision trees is taken to predict the class of a given input.
\paragraph{Gradient Boosting (XGB)} XGB refers to eXtreme Gradient Boosting. It is an implementation of boosting that supports training the model in a fast and parallelized way. Boosting is another ensemble classifier combining a number of weak learners, typically decision trees. They are trained in a sequential manner, unlike RFs, using forward stage-wise additive modeling. During the early iterations, the decision trees learned are simple. As training progresses, the classifier becomes more powerful because it is made to focus on the instances where the previous learners made errors. At the end of training, the final prediction is a weighted linear combination of the output from the individual learners \cite{friedman2001elements}.
\paragraph{Support Vector Machines (SVMs)} SVMs are supervised learning models with associated learning algorithms that analyze data used for classification and regression analysis. An SVM training algorithm essentially builds a non-probabilistic binary linear classifier (although methods such as Platt scaling \cite{platt1999probabilistic} exist to use SVM in a probabilistic classification setting). It represents each training example as a point in space, mapped such that the examples of the separate categories are divided by a clear gap that is as wide as possible (this is usually achieved by minimizing the hinge loss). New examples are then mapped into that same space and predicted to belong to a category based on which side of the gap they fall. SVMs were originally introduced to perform linear classification; however, they can efficiently perform a non-linear classification using the kernel trick \cite{cortes1995support}, implicitly mapping their inputs into high-dimensional feature spaces.
\paragraph{Multinomial Naive Bayes (MNB)} Naive Bayes classifiers are a family of simple ``probabilistic classifiers" based on applying Bayes' theorem with strong (naive) independence assumptions between the features. Under multinomial settings, the feature vectors represent the frequencies with which certain events have been generated by a multinomial $(p_{1},\dots ,p_{n})$ where $p_{i}$ is the probability that event $i$ occurs. MNB is very popular for document classification task in text \cite{kibriya2004multinomial} which too essentially is a multi-class classification problem.
\paragraph{Logistic Regression (LR)} LR is typically used for binary classification problems \cite{king2001logistic}, that is, when we have only two labels. In this work, LR is implemented in a one-vs-rest manner; six classifiers have been trained for each class and finally, we consider the class that is predicted with the highest probability.

Having trained the above classifiers, we take ensemble of the best performing classifiers and use it for comparison with the current state-of-the-art for emotion recognition on the IEMOCAP dataset.

\subsection{Deep Learning Models}
\begin{figure}[h]
	\centering
	\includegraphics[width=\linewidth]{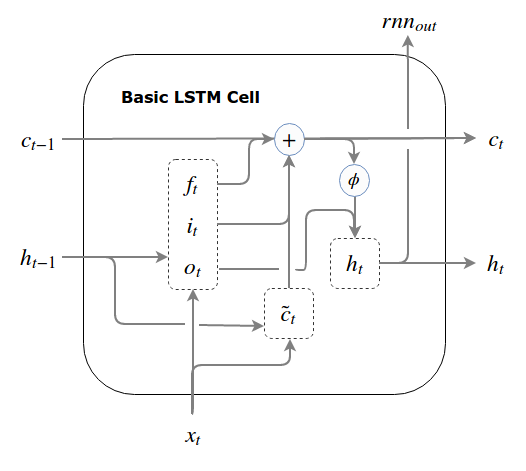}
	\caption{Visualization of an LSTM cell}
	\label{fig:lstm}
\end{figure}

In this section, we describe the deep learning models used. Typically, Deep Neural Networks (DNNs) are trained in an end-to-end fashion and they are expected to ``figure out" features completely on their own. However, training such a model can take a lot of time as well as computational resources. In order to minimize the computational overhead, we directly feed the handcrafted features as input to these models and compare their performance with the traditional end-to-end trained counterparts. In this work, we implement two types of models:
\paragraph{Multi-Layer Perceptron (MLP)} MLP belongs to a class of feed-forward neural network. It consists of at least three nodes: an input, a hidden and an output layer. All the nodes are interleaved with a non-linear activation function to stabilize the network during training time. Their expressive power increases as we increase the number of hidden layers upto a certain extent. Their non-linear nature allows them to distinguish data that is not linearly separable.
\paragraph{Long Short Term Memory (LSTM)} LSTMs \cite{hochreiter1997long} were introduced for long-range context capturing in sequences. Unlike MLP, it has feedback connections that allow it to decide what information is important and what is not. It consists of a gating mechanism and there are three types of gates: input, forget and output. Their equations are mentioned below:

\begin{equation}
    f_t = \sigma_{g}(W_fx_t + U_fh_{t-1} + b_f)
\end{equation}

\begin{equation}
    i_t = \sigma_{g}(W_ix_t + U_ih_{t-1} + b_i)
\end{equation}

\begin{equation}
    o_t = \sigma_{g}(W_ox_t + U_og_h{t-1} + b_o)
\end{equation}

\begin{equation}
    c_t = f \cdot c_{t-1} + i_t \cdot \sigma_{c}(W_cx_t + U_ch_{t-1} + b_c)
\end{equation}

\begin{equation}
    h_t = o_t \cdot \sigma_{h}(c_t)
\end{equation}

where initial values are $c_0 = 0$ and $h_0 = 0$ and $\cdot$ denotes the element-wise product, $t$ denotes the time step (each element in a sequence belongs to one time step), $x_t$ refers to the input vector to the LSTM unit, $f_t$ is the forget gate's activation vector, $i_t$ refers to the input gate's activation vector, $o_t$ refers to the output gate's activation vector, $h_t$ is the hidden state vector (which is typically used to map a vector from the feature space to a lower-dimensional latent space,) $c_t$ is the cell state vector and $W, U \text{ and } b$ are weight and bias matrices which need to be learned during training. From figure \ref{fig:lstm}, we see that an LSTM cell is able to keep track of hidden states at all time steps through the feedback mechanism.

\begin{figure}[h!]
	\centering
	\includegraphics[width=\linewidth]{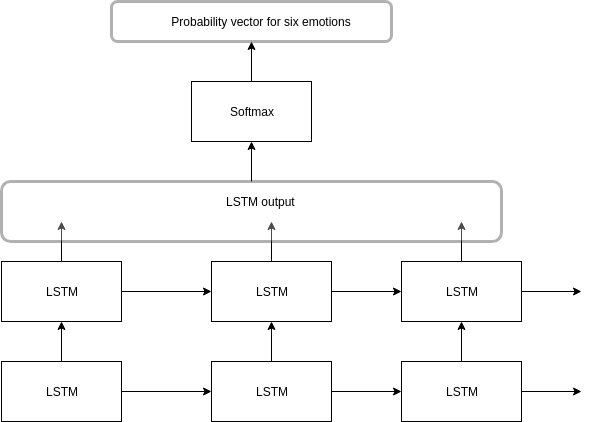}
	\caption{LSTM classifier}
	\label{fig:lstm_classifier}
\end{figure}

Figure \ref{fig:lstm_classifier} shows the network implemented in this work. We feed the feature vectors as input to the network and finally pass the output of the LSTM network through a softmax layer to get probability scores for each of the six emotion classes. Since we are using feature vectors as input, we do not need another decoder network to transform it back from hidden to output space thereby reducing network size.

\begin{figure}
    \centering
    \begin{subfigure}[ht]{\linewidth}
    	\centering
    	\includegraphics[width=\linewidth]{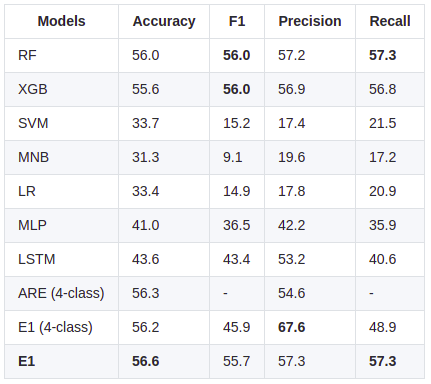}
    	\caption{Audio-only setting}
    	\label{fig:aud_res}
    \end{subfigure}

    \begin{subfigure}[ht]{\linewidth}
        \centering
    	\includegraphics[width=\linewidth]{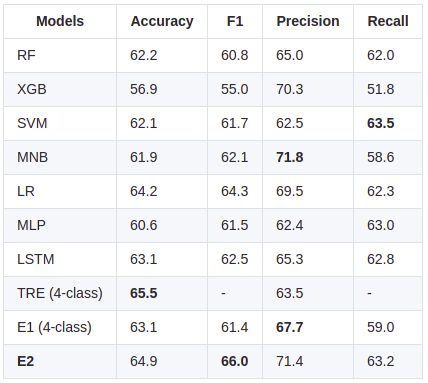}
    	\caption{Text-only setting}
    	\label{fig:tex_res}
    \end{subfigure}

    \begin{subfigure}[ht]{\linewidth}
        \centering
    	\includegraphics[width=\linewidth]{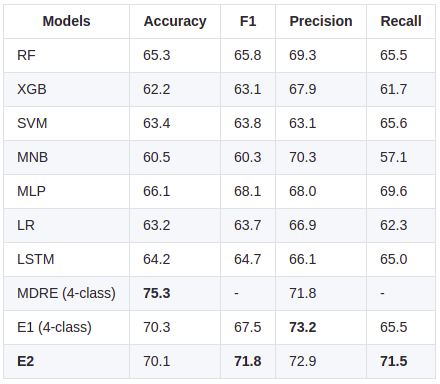}
    	\caption{Audio+Text setting}
    	\label{fig:com_res}
\end{subfigure}
    \caption{Performance of different models; E1: Ensemble (RF + XGB + MLP); E2: Ensemble (RF + XGB + MLP + MNB + LR)}
    \label{fig:performance}
\end{figure}

\begin{figure}
    \centering
    \begin{subfigure}[ht]{\linewidth}
        \centering
    	\includegraphics[width=\linewidth]{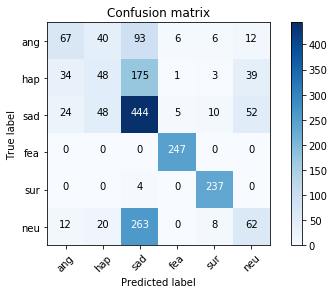}
    	\caption{E1, Audio-only setting}
    	\label{fig:cm_aud}
    \end{subfigure}

    \begin{subfigure}[ht]{\linewidth}
        \centering
    	\includegraphics[width=\linewidth]{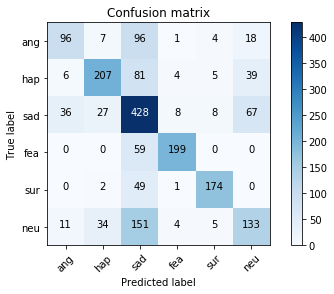}
    	\caption{E2, Text-only setting}
    	\label{fig:cm_tex}
    \end{subfigure}

    \begin{subfigure}[ht]{\linewidth}
        \centering
    	\includegraphics[width=\linewidth]{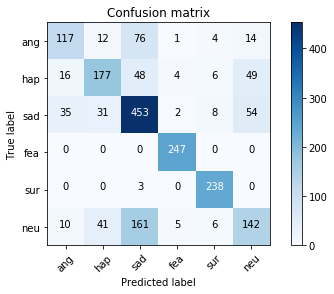}
    	\caption{E2, Audio+Text setting}
    	\label{fig:cm_com}
    \end{subfigure}
    \caption{Confusion Matrices of the our ensemble models; E1: Ensemble (RF + XGB + MLP); E2: Ensemble (RF + XGB + MLP + MNB + LR)}
    \label{fig:cm}
\end{figure}

\subsection{Experiments}
Here, we describe the three different settings we conducted our experiments in:
\begin{itemize}
    \item \textit{Audio-only:} In this setting, we train all the classifiers using only the audio feature vectors described earlier.
    \item \textit{Text-only:} In this setting, we train all the classifiers using only the text feature vectors (TFIDF vectors)
    \item \textit{Audio+Text:} In this setting, we fuse the feature vectors from the two modalities. There have been some methods proposed to fuse vectors efficiently from multiple modalities but we simply concatenate the feature vectors from audio and text to obtain the combined feature vectors. Through this experiment, we would be able to infer how much information is contained in each of the modalities and how does fusion influence the model's performance.
\end{itemize}

\subsection{Implementation Details}
In this section, we describe the implementation details adopted in this work.
\begin{itemize}
    \item We use \texttt{librosa} \cite{mcfee2015librosa}, a Python library, to process the audio files and extract features from them.
    \item We use \texttt{scikit-learn} and \texttt{xgboost} \cite{pedregosa2011scikit} \cite{dmlcsgb}, the machine learning libraries for Python, to implement all the ML classifiers (RF, XGB, SVM, MNB, and LR) and the MLP.
    \item We use \texttt{PyTorch} \cite{pytorch} to implement the LSTM classifiers described earlier.
    \item In order to regularize the hidden space of the LSTM classifiers, we use a \textit{shut-off} mechanism, called dropout \cite{srivastava2014dropout}, where a fraction of neurons are not used for final prediction. This is shown to increase the robustness of the network and prevent overfitting.
\end{itemize}

We randomly split our dataset into a train (80\%) and test (20\%) set. The same split is used for all the experiments to ensure a fair comparison. The LSTM classifiers were trained on an NVIDIA Titan X GPU for faster processing. We stop the training when we do not see any improvement in validation performance for $>$10 epochs. Here, one epoch refers to one iteration over all the training samples. Different batch sizes were used for different models. Hyperparameters for the all the models under the three experiment settings could be found in the released repository.

\subsection{Evaluation Metrics:}
In this section, we first describe the various evaluation metrics used and report results for the three experiment settings.
\paragraph{Accuracy} This refers to the percentage of test samples that are classified correctly.
\paragraph{Precision} This measure tells us out of all predictions, how many are actually present in the ground truth (a.k.a. labels). It is calculated using the formula:

\begin{equation}
    Precision = \frac{tp}{tp + fp}
\end{equation}
\paragraph{Recall} This measure tells us how many correct labels are present in the predicted output. It is calculated using the formula:

\begin{equation}
    Precision = \frac{tp}{tp + fn}
\end{equation}

Here, $tp$, $fp$, and $fn$ stand for true positive, false positive and false negative respectively. We can compute these values from the confusion matrix.

\paragraph{F-score} It is defined as the harmonic mean of precision and recall. This measure was included as accuracy is not a complete measure of a model's predictive power but F-score is since it is more normalized.

We compare our best performing models with the current state-of-the-art as mentioned in \cite{yoon2018multimodal}. They employ three types of recurrent encoders, namely, ARE, TRE and MDRE denoting Audio-, Text- and Multimodal Dual- Recurrent Encoders respectively. It is important to mention that \cite{yoon2018multimodal} only considers four emotions for classification, namely, angry, happy, sad and neutral as opposed to six in our case. In order to present a fair comparison of our method with theirs, we also run the experiments for the four classes (models with code 4-class in Figure \ref{fig:performance}).

\section{Results}
In this section, we discuss the performance of models described in Section IV.

From Figure \ref{fig:performance}, we can see that our simpler and lighter ML models either outperform or are comparable to the much heavier current state-of-the-art on this dataset. A more detailed analysis follows:

\paragraph{Audio-only results} Results are especially interesting for this setting. Performance of LSTM and ARE reveals that deep models indeed need a lot of information to learn features as the LSTM classifier trained on eight-dimensional features achieves very low accuracy as compared to the end-to-end trained ARE. However, neither of them are able to beat the lighter E1 model (Ensemble of RF, XGB and MLP) which was trained on the eight-dimensional audio feature vectors. A look at the confusion matrix (Fig. \ref{fig:cm_aud}) reveals that detecting ``neutral" or distinguishing between ``angry", ``happy" and ``sad" is the most difficult for the model.
\paragraph{Text-only results} We observe that the performance of all the models for this setting is similar. This could be attributed to the richness of TFIDF vectors known to capture word-sentence correlation. We see from the confusion matrix (Fig. \ref{fig:cm_tex}) that our text-based models are able to distinguish the six emotions fairly well along with the end-to-end trained TRE. We observe that ``sad" is the toughest for textual features to identify very clearly.

\paragraph{Audio+Text results} We see that combining audio and text features gives us a boost of $\sim$14\% for all the metrics. This is clear evidence of the strong correlation between text and speech features. Also, this is the only case when the recurrent encoders seem to perform slightly better in terms of accuracy but at the cost of precision. The lower performance of E1 maybe be attributed to the trivial fusion method (concatenation) we use as simple concatenation for an ML model would still contain a lot of modality-specific connections instead of the desired inter-modal connections. The promising result here is that combining features from both the modalities indeed helped to resolve the ambiguity observed for modality-specific models as shown in Fig. \ref{fig:cm_com}. We can say that the textual features helped in correct classification of ``angry" and ``happy" classes whereas the audio features enabled the model to detect ``sad" better.

Overall, we can conclude that our simple ML methods are very robust to have achieved comparable performance even though they are modeled to predict six-classes as opposed to four in previous works.

\subsection{Most Important Features:}
In this section, we investigate which features contribute the most during prediction in this classification task. We chose the XGB model for this study and rank the eight audio features. We see that \textit{Harmonic}, which is directly related to the excitation in signals, contributes the most. It is interesting to see that ``silence" attributing to \textit{Pause}, is almost as significant as standard deviation of the autocorrelated signal (related to pitch). The low contribution of central moments is expected as a signal is very diverse and an global/coarse feature would be unable to identify the nuances present in it.

\begin{figure}[h]
    \centering
	\includegraphics[scale=0.22]{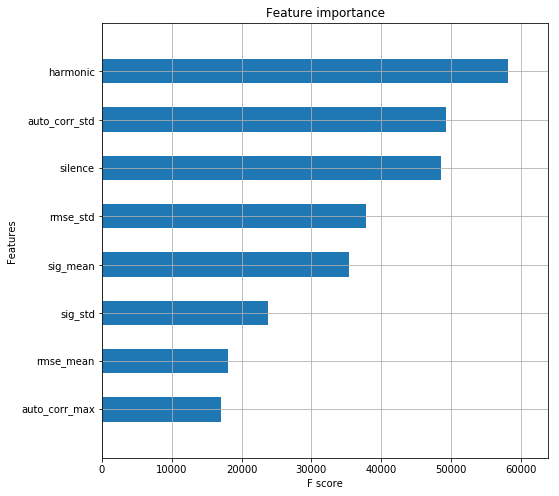}
	\caption{Most important Audio features}
	\label{fig:feat_imp}
\end{figure}

\section{Conclusion and Future Work}
In this work, we tackle the task of speech emotion recognition and study the contribution of different modalities towards ambiguity resolution on the IEMOCAP dataset. We compare, both, ML- and DL-based models and show that even lighter and more interpretable ML models can achieve performance close to DL-based models. We show that ensembling multiple ML models also lead to some improvement in the performance. We only extract a handful of time-domain features from audio signals. The audio feature-space could be made even richer if we could include some frequency-domain features too such as Mel-Frequency Cepstral Coefficients (MFCC) \cite{davis1980comparison}, Spectral Roll-off and additional time-domain features such as Zero Crossing Rate (ZCR) \cite{gouyon2000use}. Also, better fusion methods such as TFN \cite{zadeh2017tensor} and LMF \cite{liu2018efficient} could be employed for combining speech and text vectors more effectively. It would also be interesting to see the scaling in the performance of ML models v/s DL models if include more data.

\newpage

\bibliographystyle{ieeetr}
\bibliography{main}
\end{document}